\documentclass[11pt, conference]{IEEEtran}
\usepackage{color, pbox}
\usepackage{amsmath}
\DeclareMathOperator*{\argmin}{arg\,min}

\ifCLASSINFOpdf
\else
\fi
\begin{document}
%
\title{Want Answers?\\ A Reddit Inspired Study on How to Pose Questions}

\author{\IEEEauthorblockN{Danish}
\IEEEauthorblockA{Indian Institute of Science\\
Bangalore, India\\
Email: danish037@gmail.com}
\and
\IEEEauthorblockN{Yogesh Dahiya}
\IEEEauthorblockA{Indian Institute of Science\\
Bangalore, India\\
Email: yogeshd2612@gmail.com}
\and
\IEEEauthorblockN{Partha Talukdar}
\IEEEauthorblockA{Indian Institute of Science\\
Bangalore, India\\
Email: ppt@serc.iisc.in}}


%


\newcommand{\refalg}[1]{Algorithm~\ref{#1}}
\newcommand{\refeqn}[1]{Equation~\ref{#1}}
\newcommand{\reffig}[1]{Figure~\ref{#1}}
\newcommand{\reftbl}[1]{Table~\ref{#1}}
\newcommand{\refsec}[1]{Section~\ref{#1}}
\newcommand{\method}[1]{\mbox{\textsc{#1}}}

\newcommand{\simrfull}{Single Inquirer Multiple Responders}
\newcommand{\simr}{SIMR}
\newcommand{\misrfull}{Multiple Inquirers  Single Responder}
\newcommand{\misr}{MISR}

\newcommand{\reminder}[1]{}

\newcommand{\system}{SYSTEM}
\newcommand{\systemfull}{SYSTEM FULL}

\newtheorem{theorem}{Theorem}[section]
\newtheorem{claim}[theorem]{Claim}

\maketitle

\begin{abstract}
Questions form an integral part of our everyday communication, both offline and online. Getting  responses to our questions from others is fundamental to satisfying our information need and in extending our knowledge boundaries. A question  may be represented using various factors such as social, syntactic, semantic, etc. We hypothesize that these factors contribute with  varying degrees towards getting responses from others for a given question. We perform a thorough empirical study to
measure effects of these factors using a novel question and answer dataset from the website Reddit.com. To the best of our knowledge, this is the first such analysis of its kind on this important topic. We also use a sparse non-negative matrix factorization technique to automatically induce \emph{interpretable} semantic factors from the question dataset. We also document various patterns on response prediction we observe during our analysis in the data. For instance, we found that preference-probing questions are scantily answered. Our method is robust to capture such latent response factors. We hope to make our code and datasets publicly available upon publication of the paper.

\end{abstract}



%
\IEEEpeerreviewmaketitle

\section{Introduction}
\label{sec:intro}

Questions and the responses they elicit  are a ubiquitous and fundamental part of 
our everyday communication.  Through such Questions and Answers (QA), we quench our curiosities, clarify doubts, validate our ideas, and seek advice, among others. It has been established that questions form an integral part in our quest to extend our knowledge boundaries \cite{sammut1986learning}. It has also been observed that useful responses correspond to \textit{good} questions \cite{agichtein2008finding}. This raises the following challenge:  \textit{what factors constitute a good question which is more likely to elicit a response}? \reminder{PPT: maybe we should include an example}

Importance of asking right questions in specific settings have been previously explored, e.g., in classroom \cite{king1994guiding}, and in corporate environment \cite{ross2009ask}. However, most of these studies either had no empirical evaluation at all or otherwise consisted of very small samples.

Along with the growth of the World Wide Web (WWW), many large online QA sites, such as Yahoo Answers, Stack Overflow, Quora, etc., have been successful in connecting responders to inquirers who post questions on these sites. Such online QA forums may be categorized as  {\bf \simrfull{} (\simr{})}, where a question from a single user may be  responded to by multiple other responders. Prior research have used datasets from these sites to analyze which response to a question is most likely to be selected as the best response \cite{adamic2008knowledge}  \reminder{PPT: additional citations}. However, analyzing factors of a question which are  likely to elicit a response has been outside the scope of such prior work. 

To address these shortcomings, in this paper we present an empirical analysis to determine factors of a question which  are more likely to elicit a response. We make use of the IAmA subreddit of the popular Internet website Reddit.com. In each discussion thread of this online forum, a celebrity answers questions submitted by anonymous users. Thus, dataset from this subreddit may be categorized as {\bf \misrfull{} (\misr{})}. Such \misr{} datasets provide an ideal starting point to identify response-eliciting factors of a question, as the undesirable confounds produced due to the presence of multiple responders in \simr{} datasets are not present in such \misr{} datasets.

We make the following contribution:

\begin{itemize}
	\item We address the important problem of automatically identifying response-eliciting factors of a question. We explore effectiveness of various factors, viz., orthographic, temporal, syntactic and also semantic of the question. To the best of our knowledge, this is the first such analysis of its kind.
	\item We make use of a novel dataset, questions and responses from the IAmA subreddit of reddit.com. This MISR dataset provides additional benefits compared to \simr{} datasets which have been explored in previous related research.
	\item We provide a sparse, non-negative matrix factorization-based framework to automatically induce semantic factors of a question collection. Through extensive experiments on real datasets, we demonstrate that such factorization-based technique results in significantly more interpretable factors compared to standard topic modeling techniques, such as Latent Dirichlet Allocation (LDA). 
	\item We hope to make all the code and datasets used in the paper publicly available upon publication of the paper.
\end{itemize}


\section{Related Work}
\label{sec:related}

Studies on questioning techniques date back to Socrates \cite{paul2007critical,carey2004socratic}, who encouraged a systematic, disciplined, and deep questioning of fundamental concepts, theories, issues and problems. Socratic questioning is widely adopted in education and psychotherapy. Under the Socratic Questioning scheme \cite{paul2006thinker}, questions are grouped as follows:\\ 
i) Clarifying questions, ones seeking further explanation, 
ii) Challenging the assumptions, questions that challenge the constraints, 
iii) Argument based questions, ones that reason behind the underlying theory or seek evidence, 
iv) Alternate viewpoints, questions that analyze the given scenario with an altogether different perspective,
v) Implication and Consequence based questions

Since Socrates, many different taxonomies have been discussed. Bloom's revised taxonomy given by Krathwohl \cite{anderson2001taxonomy} is based upon dividing questions into levels such that the amount of mental activity required to respond increases after each level. Their categories are --- remembering, understanding, applying, analyzing, evaluating and creating. Nam et al.\cite{nam2009questions} group questions into Factual, Procedural, Opinion-oriented, Task-oriented and Advice related categories.

Role of Socratic Techniques in thinking, teaching and learning has also been explored \cite{elder1998role}. Hypothetical questions too have been studied independently and have been found to foster creativity \cite{newman2000hypotheticals}. While there has been considerable thought given over such demarcations and question formulation techniques, none of them are supported by any large datasets as most of the experiments were performed in a typical classroom sized setting.

Unlike the mentioned qualitative analysis, Whittaker et al.\cite{whittaker2003dynamics} adopted a data-centric approach and uncovered the general demographic patterns among large samples of Usenet newsgroups. Some amount of research is also done on different QA forums like Yahoo! answers, e.g., \cite{adamic2008knowledge} have proposed solutions to predict whether a particular answer will be chosen best by the inquirer or not. 

Although the aforementioned taxonomies are helpful in understanding general questioning paradigms, we are more curious about the qualities of a question that are more likely to generate a response. To the best of our knowledge, there have been no attempts to study questions with the objective of maximizing response rate.  

Variety of interesting questions have been studied using the Reddit conversation network. It has been used to understand how people react to online discussions\cite{jaech2015talking}, and to model the most reportable events in stories\cite{ouyang2015modeling}. Domestic abuse analysis in \cite{ray4183analysis} also was based upon Reddit. 

An empirical case study to understand  factors underlying successful favor requests online were studied in \cite{althoff2014ask}. Like in that paper, we also make use of a subreddit as our primary dataset. Even though the setting explored in \cite{althoff2014ask} is different than ours, this is probably the closest paper in motivation and spirit.

Unlike other analysis on online forums including Reddit and Yahoo! answers, our dataset is unique as it falls into \misr{} category, where there is just one responder but multiple inquirers. To the best of our knowledge, this is a first attempt to understand any such dataset.


\section{Dataset}
\label{sec:dataset}

Reddit is the 26th most popular website, with about 36 million user accounts. It also comprises of over 9,000 subreddits which are sub-forums within reddit, these subreddits are focussed towards specific topics. Subreddits span diverse categories like News, Sports, Machine Learning etc. Reddit is also a home of subreddits like: ELIF (Explain like I'm five), TIL (Today I learnt), AMA(Ask Me Anything) etc. 

Various celebrities and noteworthy personalities have used reddit as a means to interact with the popular internet crowd, such conversations fall under the Ask-Me-Anything and its variant subreddits. IAmA, AMA and casualama are 3 of the popular Ask-Me-Anything variants. IAmA is reserved for distinguished personalities, with an exception for people who have a truly interesting and unique event to take questions about. 
The other two AMA's are open to a more wider audience for sharing their life events and allowing other reddit users to ask questions related to those events.

IAmA's is one of the most popular subreddits that has featured notable politicians, actors, directors, authors, businessmen, athletes and musicians. IAmA posts gain a lot of attention, and thousands of questions are asked in each IAmA post. But owing to time constraints, not all questions are answered. This gives us a good ground to understand and analyze what gets answered and what not.

In particular, we study four popular categories of celebrities --- actors, authors, directors and politicians. In each category we analyse the top 50 upvoted posts, which aggregate over 110,000 questions, and the average reply rate is 10.16\%. Since some questions arrive after the celebrity has moved out of the conversation, we ignore all the questions after the last successfully answered question.
Reddit allows for threaded conversations, where users can comment over other comments. But to avoid any bias from the discourse of the comments in such threads, we ignore questions in deep threaded conversations and constrain ourselves to questions posted at the topmost level only. Since some comments also get posted at the topmost level, we only consider comments that have a question mark in them. 
\reftbl{tbl:datasets} throws some light over the statistics about the questions we considered as a part of our study.   

\begin{table}[t]
\begin{center}
\begin{tabular}{||c|p{1.5cm}|p{1.5cm}|p{1.5cm}||}
\hline
Domain & Questions Asked & Questions Replied & Response Rate \\
\hline\hline
Actor & 58859 & 3060 & 5.19 \\
\hline
Author & 21295 & 3752 & 17.61 \\
\hline
Politician & 13866 & 1914 & 13.8 \\
\hline
Director & 24196 & 3295 & 13.61 \\
\hline
Total & 118216 & 12021 & 10.16 \\
\hline
\end{tabular}
\caption{\label{tbl:datasets}Reddit IAmA datasets from four domains used in the experiments in this paper. See \refsec{sec:dataset} for further details}
\end{center}
\end{table}

\section{Success Factors of Questions}
\label{sec:method}

In this section, we study various factors of questions that can result in healthy response rates. The factors we consider  range from orthographic, temporal, social, and syntactical, to semantic aspects. 

\subsection{Orthographic Factors}

{\bf Length}: Do short questions win over their longer variants, as the responder may not be interested in comprehending and then answering long questions? Or, are longer questions better as they offer more context? Are shorter and crisper questions more direct and focused and have a better chance at getting answered? We analyze the impact of question on response rate to answer the aforementioned question. 

\subsection{Temporal Factors}

{\bf Time of Question}: Does the time of asking question play any role in determining the response rate? We hypothesize that questions that are asked early on have far less competing questions and hence should have better  chances of soliciting response.

We capture temporal information in two ways: (1) we note the fraction of \emph{questions answered} in the IAmA before a given question is posted as an estimate of the time of question; (2) we use the fraction of \emph{time elapsed} in the IAmA as another indicator of the time of the question. In most cases, we see that the time features complement each other.

\subsection{Social Factors}
\label{sec:social-factors}

{\bf Politeness}: Are polite questions more likely to generate a response? Or, is it the case that the default level of politeness expressed in the IAmA dataset already sufficient, and hence any additional politeness in the question is unlikely to positively affect response rate? 

Politeness has been actively explored in the recent past in a variety of others research settings\cite{tsang2006brief}\cite{bartlett2006gratitude}. We employ the model introduced by Danescu-Niculescu-Mizil et al.  \cite{danescu2014computational} to measure politeness level of questions. This model bases its politeness score on the occurrences of greetings, apologies and hedges in the question.

\subsection{Syntactic Factors}
\label{sec:syn_features}
{\bf Syntactic}: We ask whether questions that are simply formulated have better chances of getting answered? Syntactic features, such as parse tree depth, verb phrase depth, and their ratios \cite{klein2003accurate}, etc., have been used in past research as proxies for sentence complexity. In fact, such features have also been recently used to study syntactic complexity of reddit comments \cite{ouyang2015modeling}. After generating constituent parse trees from the Stanford Corenlp package \cite{manning2014stanford}, we employ 16 such features to capture the essence of syntactic complexity in a given question. 

We look at a few simple and a few complex sentences from the IAmA by President Barack Obama in \reftbl{tbl:syntax}. We demonstrate how our features capture the varied levels of complexity. Since there can be various sentences and sub-questions in a given question, we calculate the average, maximum and minimum values of parse tree depths and verb phrase depths. 
It is because of such statistical aggregation techniques that we end up with 16 syntax features, but the basis of these features rest upon  --- parse tree of the sentence, verb phrase subtree and their ratios. 

\begin{table*}[t]
\begin{center}
	\begin{tabular}{|c|c|c|c|c|}
		\hline
		Sentence & Depth of Sentence & \#Verb Phrases & Max Verb Phrase & Ratio of Verb Phrase\\
							 & Parse Tree & & Depth & and Sentence Depth\\
		\hline
		Who's your favourite Basketball player? & 2 & 0 & 0 & 0.0\\
		\hline.
		What's the recipe for the White House's beer? & 6 & 0 & 0 & 0.0\\
		\hline
		How can we help to increase the standard of debate in this country & & & & \\ 
		so that it's not simple soundbytes and broad generalizations & 7& 3 & 6 & 0.86\\
		of complicated stuff & & & & \\
		\hline
		Mr. President - What issues, if any,do you agree with Mitt Romney that & 11 & 4 & 9 & 0.81\\
		are not commonly endorsed by the majority of the Democratic Party? & & & &\\
		\hline 

	\end{tabular}
	\caption{\label{tbl:syntax}A few example sentences from President Obama's IAmA and their corresponding syntax features. See \refsec{sec:syn_features} for details}
\end{center}	
\end{table*}


\subsection{Forum Factors}
\label{sec:forum_factors}

{\bf Redundancy}: Is a question which is very similar to already asked (or answered) questions in a given IAmA forum less likely to get a response? We think that is indeed the case and include factors in our analysis to account for question redundancy. Consider the questions in \reftbl{tbl:examples} asked to a popular Chef. \\
\begin{table}[t]
\begin{center}
	\begin{tabular}{|c|}
		\hline
		What's your favorite Middle Eastern Dish? \\
		\hline
		What's your favourite dish to prepare?\\
		\hline
		What's your favourite French meal? \\
		\hline
		What dish changed your life? \\
		\hline
		Hey Mr Ramsay, I'm a big fan of your Kitchen Nightmare series \\ What is your favourite dish to cook at home?\\
		\hline
	\end{tabular}
	\caption{\label{tbl:examples}A few examples of redundant questions asked to a Chef. See \refsec{sec:forum_factors} for details}
\end{center}
\end{table}


As the first few questions were not answered in the series of the above mentioned questions, it is nearly certain that the responder is not interesting in any such questions. By accounting for redundancy we hope to tackle similar and frequent scenarios. 

We estimated the redundancy score of a given question as the maximum similarity score achieved with any of the other questions previously asked in the same IAmA. 

{\bf Relevance}: For each IAmA, the responder usually posts a description to set the tone of the IAmA. We ask whether questions which are more aligned to the posted description more likely to receive a response? The posted descriptions usually carry information about the celebrity responder's current affiliation and engagements, and hence the hypothesis is that questions which are in line with such descriptions should outweigh other questions. In other words, relevant questions should attract more responses from the responder. 

For both the relevance and redundancy factors, we came up with our own novel extension of Jaccard Similarity to account for sentence similarities. For two given sets A and B, the Jaccard Similarity is given by
\begin{equation}
	J(A, B) = \frac{|A \cap  B|}{|A \cup B|}
\end{equation}

For our case, let A and B be sets of words corresponding to the two questions to be compared. Strictly, \(A \cap B\) would translate to the count of the words matched across the sets of A and B. But consider the following two sentences: \\
\textit{-- How far is your workplace from your house?} \\
\textit{-- How far is your office from your home?}\\

With the strict definition, we would not be able to capture that the two sentences are completely similar, for all practical purposes. Hence we consider the Glove embeddings\cite{pennington2014glove}, and synset hierarchies to extend the scope of our matching. Two words are considered same, if (1) the two words are synonyms to each other and (2) if one word lies in top-K nearest neighbours of the other word in Glove embedding space. 

This technique helps us to capture similarity of pairs like $<$home, house$>$ and $<$office, workplace$>$ and hence helps us better estimate the similarity of two sentences.

\subsection{Semantic Factors}
\label{sec:sem_features}
The factors described so far consider various aspects of the questions being analyzed. However, none of them explicitly look at the semantic content of the question and perform analysis based on the semantic type of the question. For example, given questions of the following form posed to actors, \textit{"what is your favorite movie?"}, \textit{"what is your favorite book?"}, etc.  we would like to group all such preference-probing questions into one category and then determine the response
rate for such types of questions from actor responders. However, such categorization of questions are not readily available as we only have the list of question, and no additional annotation on top of them. 

Ideally, we would like to discover such categorical structure in the data automatically. Topic modeling techniques such as Latent Dirichlet Allocation (LDA) \cite{blei2003latent} may be employed to discover such latent structure in the question dataset. Given a set of questions, such techniques will induce topics as probability distribution over words. Ultimately, each question is going to be represented in terms of such induced topics. We note that interpretability, i.e., coherence among questions which share a given topic with high weights, is of paramount importance here as all subsequent response-rate analysis are going to be hinged on the label or meaning of each topic. Unfortunately, as we shall see in \refsec{sec:expts}, topics induced by LDA   don't achieve the desired level of interpretability.

To overcome this limitation, we explore other latent factorization methods. Recently, Non-Negative Sparse Embedding (NNSE) \cite{murphy2012learning,fyshe2015compositional} has been proposed which tends to induce effective as well as interpretable embeddings. In order to apply NNSE to our question dataset, we first represent the data as a co-occurrence matrix $X$ where rows correspond to questions and columns correspond to words. Each question is additionally augmented with
word sense-restricted synsets from Wordnet. The effect after the synset extension from Wordnet can be seen in \reftbl{tbl:extension}  This extended co-occurrence matrix $X$ is usually of very high dimension (e.g., 100k x 1m). We first reduce dimensionality of the matrix using sparse SVD. The number of dimensions in the SVD space is selected based on knee-plot analysis of eigenvalues obtained during SVD decomposition. The rank $r$  approximation $X_{n \times r}$ obtained from SVD
is then factorized into two matrices using NNSE, which minimize the following objective.
\begin{equation*}
\argmin\limits_{A,D} \frac{1}{2} \sum\limits_{i = 1}^{n} \parallel X_{i,:} - A_{i,:} \times D \parallel^2 
\end{equation*}
\begin{align*}
st &: D_{i,:}D_{i,:}^T \leq 1, \forall 1 \leq i \leq k \\
    &~A_{i,j} \geq 0, 1 \leq i \leq n, 1 \leq k \\
	&~\parallel A_{i:} \parallel_1  \leq  \lambda_1, 1 \leq i \leq n \\
\end{align*}
where $n$ is the number of questions, and $k$ is the resulting number of latent factors induced by NNSE. We note that NNSE imposes non-negativity and sparsity penalty on the rows of matrix $A$. Though the objective represents a non-convex system, but when we solve for $A$ with a fixed $D$ (and vice versa) the loss function is convex. In such scenarios Alternating Minimization has been established to converge to a local optima \cite{mairal2010online, murphy2012learning}. 
The solution for $A$ is found with LARS implementation\cite{efron2004least} of LASSO regression with non-negativity constrains; and $D$ is found via gradient descent methods. The SPAMS package may be used for this optimization \cite{bach2010sparse}. At the end of this process, $A_{i,j}$ represents the membership weight of question $i$ belonging to latent factor $j$.

\begin{table}[t]
\begin{center}
	\begin{tabular}{|c|c|c|c|c|}
	\hline
	Tags & \#Questions & \#co-occurrence & \#co-occurrence & Factor \\
	& &entries & after & increase \\
	\hline
	Author & 21295 & 349288 & 679476 & 1.94\\
	\hline 
	Actor & 58859 & 701702 & 1368703 & 1.95\\
	\hline
	Politics & 13866 & 228820 & 438089 & 1.91\\
	\hline
	Director & 24196 & 344176 & 658226 & 1.91\\
	\hline	
\end{tabular}	
\caption{\label{tbl:extension}Effect of extension using Wordnet synsets on the co-occurence matrix. See \refsec{sec:sem_features}}
\end{center}
\end{table}

\section{Experiments}
\label{sec:expts}

\begin{table}[t]
\centering
\begin{tabular}{|c|p{1.4cm}|p{1.4cm}|c|}
 \hline
 Domain & Temporal Factor Feature 1 & Temporal Factor Feature 2 & Redundancy\\
 \hline
 Author &  89.57 & 83.24 & 56.92\\
 Politician & 84.29 & 115.91 & 63.06\\
 Actor & 189.81 & 218.37 & 141.23\\
 Director & 63.92 & 85.37 & 47.42\\
 \hline
\end{tabular}
\caption{\label{tab:AP}Average Precision (AP) gains for temporal and redundancy factors over a random baseline. See \refsec{sec:temporal_feats} for details.}
\end{table}

\begin{table*}[t]
\centering
\begin{tabular}{|l|c|c|c|c||c|}
 \hline
 Feature (Factor) & Actor &  Author & Politician & Director & Average \\
 \hline
 Random Baseline & .50 &. 50 & .50 & .50 & 0.50 \\
 \hline
 Length (Orthographic) & .48 & .49 & .54 & .52 & 0.51 \\
 Syntactic & .53 & .52 & .53 & .50 & 0.52 \\
 Syntactic + Length & .54 & .52 & .53 & .49 & 0.52 \\
 Temporal & .66 & .67 & .67 & .60 & 0.65 \\
 Redundancy (Forum) & .71 & .65 & .64 & .62 & 0.66 \\
 Relevance (Forum)  & .49 & .51 & .58 & .51 & 0.52 \\
 Politeness (Social) & .48 & .52 & .54 & .52 & 0.52 \\
 Politeness + Relevance + Redundancy & .72 & .66 & .69 & .64 & 0.68 \\
 Unigram & .68 & .66 & .64 & .61 & 0.65 \\
 Temporal + Syntax + Unigram & .75 & .72 & .73 & .66 & 0.72 \\
 Temporal + Politeness + Relevance + Redundancy & .73 & .70 & .73 & .64 & 0.70 \\
 Temporal + Politeness + Relevance + Redundancy + Syntax & .74 & .70 & .73 & .64 & 0.70 \\
 \hline
\end{tabular}
\caption{\label{tab:AUC}ROC AUC values for a regularized logistic regression classifier using different features in various domains. For reference, performance of a random baseline is also shown. Apart from length, all other features improve performance over the random baseline. See \refsec{sec:predict_response}. See \refsec{sec:expts} for details.}
\end{table*}

%
%

\begin{table*}[t]
\centering
\begin{tabular}{|c|p{0.8cm}|p{15cm}|}
 \hline
 {\bf Method} & {\bf Latent Factor} \# & {\bf Top Two Questions in the Latent Factor} \\
 \hline
  LDA & 1 & \textit{-- Do you think that if you lived in an urban environment when these stories came to you, you might have {\bf written about} rats or pigeons?} \\
   & & \textit{-- to what extent should historical analysis of religious figures impact the way people practice {\bf faith}? Or do you feel that the events of history are independent from the values of modern religion?} \\
 \cline{2-3}
   & 2 & \textit{-- chuck, i used to own a book by you, but i don't remember the title nor have i seen it anywhere else or included in your bibliography; {\bf did you coin the term trustafarian} .... }\\
 & & \textit{-- do you {\bf intend/hope} for the romans to speak latin with subtitles in the movie? or english?} \\
\cline{2-3}
  & 3 & \textit{-- How did everyone else in the fox news studio {\bf treat} you? were they hostile, friendly, indifferent, etc?}\\
  & & \textit{-- I {\bf love your writings}, I have read fight club, survivor, and damned and now can not wait to read doomed. My question, do you ever read neil gaiman?} \\
 \cline{2-3}
  & 4 & \textit{-- why should i {\bf care} about your hyper privilege?}\\
  & & \textit{-- I heard that you {\bf smoke} 82 blunts a day. Is that true?} \\
\hline \hline
 NNSE & 1 & \textit{-- i will ask this - Is there any {\bf advice} that you wish someone wouldve given your parents that would have smoothed out some of those painful but lets be honest here funny experiences for you} \\
  & & \textit{-- im only 21 and havent really {\bf published anything} in places people have actually heard of. I've been submitting to some lit journals thoughheres hoping writing is what I want to do for a living...} \\
 \cline{2-3}
  & 2 & \textit{-- thinking of updating the site, I preorderd book the first I heard about it it says it will be delivered on oct 29. I'm excited. {\bf I' m excited for you} I'm excited by so many things ...} \\
  & & \textit{-- Hi Robert, {\bf love your work}. Thank you I'm curious how you feel about the modern binge style of consuming tv shows and comics. I really love walking dead in both forms but I...} \\
  \cline{2-3}
  & 3 & \textit{-- ohn {\bf I'm a big fan} first off I'm a teenage guy whos read most of your books and while they do involve love I wouldnt say your books are love stories ...} \\
  & & \textit{-- Hi john whats your favourite question to be asked, Yes, I'm canadian and is there anything that youre hoping {\bf people would ask big fan of your work} both in print and on youtube} \\
 \cline{2-3}
  & 4 & \textit{-- hi steve im a constant reader like the rest of us thanks for everything im reading firestarter for the first time and {\bf love it} 1 will we ever see you work with george romero again 2 please ...} \\
  & & \textit{-- mr king {\bf I love your books} your horror fiction is fantastic I love to read it before i go to sleep. I hear youre very strict with yourself about the amount of work you put....} \\
\hline
\end{tabular}
\caption{\label{tab:lda-vs-nnse}Four randomly selected latent factors induced each by LDA and NNSE, and top ranking questions in each such factor. The main perceived theme of each question is highlighted in bold manually. We find that the factors induced by NNSE are usually much more interpretable compared to LDA. Because of this interpretability, we use semantic factors induced by NNSE for all experiments in the paper. See \refsec{sec:sem_factors} for details.}
\end{table*}

\begin{table*}[t]
\centering
\begin{tabular}{|p{1.5cm}|p{1.5cm}|p{2.8cm}|p{10cm}|}
\hline
 {\bf Domain (Overall Response Rate)} & {\bf Latent Factor \# (Response Rate)} & {\bf Sample Frequent n-grams of Questions in Latent Factor} & {\bf Top Ranking Questions in Latent Factor} \\
 \hline
 Actor (5.19\%) & 524 (15.88\%) 
 	& \textit{huge fan, loved movies, really love} & 
 		\textit{-- i m a huge fan of your cooking and have been watching you on television since i was a child so id love your input on a couple things ...} \\
 	& & & \textit{-- just like to start off by saying i love the show ... have you been involved with any popular shows or films?} \\ \cline{2-4}
  & 297 (13.79\%) 
  & \textit{story behind, behind the scenes} & \textit{-- i heard that you got a concussion and had to go the er while shooting one of the seasons whats the story behind that ...} \\
  	& & & \textit{-- hey arnold whats the story behind this picture}
  	\\ \cline{2-4}
  & 880 (0.0\%) & \textit{real life} & \textit{-- have you started saying bitch more in real life since the show started} \\
  & & & \textit{-- do you say bitch as much in real life.} \\ \cline{2-4}
  & 852 & \textit{favorite actor, }  & \textit{-- what role was your favorite to play and why} \\
  & (1.09\%) & \textit{favorite actress, favorite play} & \textit{-- hey bryan just wanted to say youre an awesome actor and i was curious what your favorite breakfast cereal} \\
 \hline \hline
 Politician (13.8\%) & \pbox{1.5cm}{927 (31.5\%)} & \pbox{2.8cm}{\textit{money, campaign, influence money, hard earned}} & \pbox{10cm}{\textit{-- has your campaign accepted any money from corporate donors if so which ones and will their contributions affect your decisions} \\ \textit{-- what about campaign money? are you running this campaign without anything to fund it?}} \\  \cline{2-4}
  & \pbox{1.5cm}{567 (28.3\%)} & \pbox{2.8cm}{\textit{issue, matter, think, social issues, net neutrality}} & \pbox{10cm}{\textit{-- what in your opinion is the most pressing issue facing the uk at the present time?} \\  \textit{-- ... david cameron himself wants to confront the european court of human rights so id like to know your take on this as well as the underlying issues ...}} \\  \cline{2-4}
  & 304 (2.43\%) & \textit{pay, wage, tax, job, minimum wage} & \pbox{10cm}{\textit{-- do you not worry that a ten pounds minimum wage would crush independent businesses and severely increase mass unemployment} \\ \textit{-- what are your thoughts on the proposals on minimum wage to 8?}} \\  \cline{2-4}
  & 567 (3.57\%) & \textit{movie, film, estate} & \pbox{10cm}{\textit{-- what do you think about the movie lego?} \\  
\textit{-- have you seen the movie ...?}} \\
 \hline \hline
 \pbox{1.5cm}{Author (17.61\%)} & \pbox{1.5cm}{742 (36.53\%)} & \pbox{2.8cm}{\textit{writing stories, advice, aspiring, approach writing}} & \pbox{10cm}{\textit{-- i am a somewhat aspiring author i write a lot on writing prompts and people there have gotten to know me a bit i am currently working on a book based on a writing prompt ... 
 	} \\ \textit{-- my question is how did your following on reddit help you get a publisher on board ...}} \\ \cline{2-4}
  & \pbox{1.5cm}{136 (34.61\%)} & \pbox{2.8cm}{\textit{idea, thought experiment, mind}} & \pbox{10cm}{\textit{-- what made you come up with that idea and how do you come up with ideas in general for your stories?} \\ \textit{-- what are your thoughts regarding the 2012 mayan prophecy?}} \\ \cline{2-4}
  & \pbox{1.5cm}{4 (7.14\%)} & \pbox{2.8cm}{\textit{inspired, inspires, work, write}} & \pbox{10cm}{\textit{-- hi john im a great fan having read all of your books bar looking for alaska may i ask what inspired you to write paper towns?} \\ \textit{-- ... i was wondering what inspired you to write what made you decide to write suspenseful novels ...}} \\ \cline{2-4}
  & \pbox{1.5cm}{118 (7.31\%)} & \pbox{2.8cm}{\textit{favorite, favourite, book, author, read}} & \pbox{10cm}{\textit{-- who is your favorite author. do you ever read your own books if so which one is your personal favorite?} \\ \textit{-- ... also out of curiosity who is your favorite author a very original question i know ?}} \\ \cline{2-4}
 \hline
\end{tabular}
\caption{\label{tab:sem_factors}Automatically induced latent semantic factors with highest and lowest  response rates in multiple domains are shown. Base response rate for the domain, and the response rate for each factor is shown in brackets. Top ranking questions in each latent factor along with the most frequent n-grams in questions belonging to the particular latent factor are also shown. We point out the interpretable nature of each semantic factor (based on high-ranking questions associated with it), which allows us to draw sample conclusion as follows: while actors are unwilling to answer questions relating to their favorites or real life, authors are more willing to answer questions relating to  supporting aspiring new authors. Ability to discover such insights using an automated process and a novel dataset is the main contribution of the paper. Please see \refsec{sec:sem_factors} for details.}
\end{table*}

%

In this section, we evaluate impact of various factors discussed in \refsec{sec:method} on response rate of questions from different domains.


\subsection{Is Response Rate Predictable?}
\label{sec:predict_response}

{\bf Datasets}:  We experiment with four popular domains --- actors, authors, director and politicians. These domains covered more than 110,000 questions, and only about 10\% of them generated a response. Statistics of the IAmA datasets are presented in \reftbl{tbl:datasets}.

{\bf Metric \& Classifier}: In order to measure response rate predictive power of a subset of factors, we train a $L_2$ and $L_1$ regularized (i.e., elastic net) classifier using only those subset of factors. Hyperparameters of the classifier is tuned using over a development set using grid search. We use area under the receiver operating characteristics curve (ROC AUC) of the classifier on held out test data as our metric. This metric essentially measures how well the classifier ranks a randomly chosen positive question over a randomly chosen negative question. Please note that the dataset is highly skewed with significantly more negative questions than positive ones. This measure provides a balanced metric while accounting for the data skew.

{\bf Baselines}: To evaluate the strength and decisiveness of our probable factors, we test our system against the random and bag-of-words (BoW) baselines. In the Random Baseline, each question is randomly given one of the two labels --- answered or not answered. 

The bag of words model comprises of each and every word in the vocabulary as a feature, hence aggregating up to thousands of features for every questions. Due to the large number of features, this Unigram model performs reasonably well, but it doesn't help us in answering our general question of --- \textit{Which factors help a question get answered?} -- because the unigram features don't generalize to the factors that we are interested in evaluating.  

Experimental results comparing performance of the classifier with different features on multiple datasets are presented in \reftbl{tab:AUC}. Based on this table, we  discuss predictive capabilities of various factors below. Please refer to \refsec{sec:method} for description of the factors and how we computed them.

\subsubsection{{\bf Orthographic Factors}} From \reftbl{tab:AUC}, we observe that the length of the questions (measured in terms of numbers of tokens in the question), the only orthographic factor feature we considered, plays practically no role in influencing response rate. This is evident from the fact that the classifier with length as the only feature achieves AUC of 0.51 on average across all four domains compared to AUC of 0.5 of the random classifier.

\subsubsection{{\bf Syntactic Factors}}
From \reftbl{tab:AUC}, we clearly see that syntax-based   features add very little little predictive power to the classifier (0.52 vs 0.50 of random). Though our syntax features are rigorous enough to capture the nuances of complexity (e.g., see \reftbl{tbl:syntax}), but the responses to questions don't heavily depend on the complexity of the sentence. We observed that combining syntax with orthographic features also didn't increase predictive power.

\subsubsection{{\bf Temporal Factors}}
\label{sec:temporal_feats}

We find that temporal features play a significant role in the response rate. This is evident from \reftbl{tab:AUC} where the classifier with temporal factor features achieves a significantly higher AUC score of 0.66 compared to random 0.5. As we had hypothesized earlier, questions that are asked early tend to be replied more often than others.

In addition to classifier's AUC score, we measured effect of temporal factors using Alternative Precision (AP) as well. For questions in a given domain, AP is computed over two ordering of the questions in that domain: (1) ordering of all questions based on the value of the temporal factor features; and (2) randomly shuffled question sequences. Percentage AP gains of the feature-based ranking over the random ranking (AP averaged over thousand trials) are summarized in \reftbl{tab:AP}. From this, we observe the clear trend that temporal factor features significantly aid in response prediction, sometimes with gains as high as 218\%. We think that the responder is initially exposed to far lesser number of questions compared to a situation in the middle or towards the end of the IAmA when the number of questions demanding his or her attention are huge.


\subsubsection{{\bf Forum Factors}}

{\bf Redundancy}: Our dataset consists of prominent celebrities, and they gain undeniably high attention among Reddit users. Due to large participation, the number of similar questions is high, as many users wish to know similar facts, preferences, likings and happenings. Redundancy comes out as one of the most promising factors in understanding questions that get answered. Examples of a few redundant questions are shown in \refsec{sec:forum_factors}. 

The original, and genuine questions, which are identified by our redundant factor feature, are heavily preferred over questions that are redundant and stale. This is established by the fact the classifier which accounts for redundancy achieves a significantly higher AUC score of 0.66 compared to the random baseline.

{\bf Relevance}: Relevance of the question, with the post description by the celebrity responder, show only faint signals with the response rate. The description given by the celebrities is usually very short to capture the variety of questions. Hence we don't see any meaningful dependencies between relevance and response rate (0.52 AUC).

Overall, with all the three forum features included, the classifier achieves an AUC score of 0.68. 

\subsubsection{{\bf Politeness}}
Politeness, a seemingly important cue for demystifying question qualities, surprisingly, didn't come out as a strong predictor of response rate. In \reftbl{tab:AUC} the classifier with politeness forum factor feature achieves an AUC score of only 0.52. We have observed that the Reddit culture is very informal, frank and open. Hence, making requests extra polite might not help while framing questions in such scenarios. Of all domains, politeness is most important in the case of prominent politicians. 
 
\subsubsection{{\bf Unigram}}

In addition to the factors mentioned above, we also experimented with the bag-of-words-based unigram model. As mentioned previously, in this case, each token of the question was added a feature. From \reftbl{tab:AUC}, we observe that the unigram model achieves an AUC of 0.68 which is significantly better than the random baseline of 0.5. However, the Unigram model uses 13704 features (averaged across all four domains). It is encouraging to note that performance of this Unigram model with thousands of features is superseded by the classifier using only 4 form factor features (AUC 0.65 vs 0.68) in the response prediction task.

\subsection{Do Induced Semantic Factors Help Discover Response Trends?}
\label{sec:sem_factors}

So far, we have tried to handcraft the seemingly most important factors but we can never account for patterns other than what we are looking for. In any large dataset as ours, creating an exhaustive set that can capture all such factors is humanly impossible. Also for each factor, we need to train a system that can well detect and measure it in an unknown question. In such scenarios, the need to automatically discover latent dimensions is essential. As mentioned in \refsec{sec:sem_features}, we use LDA and NNSE to induce semantic factors present in the question dataset. First we shall present comparisons between interpretability of factors induced by these two methods. Subsequently, we shall measure the response predictive power of these induced semantic factors. 

{\bf LDA vs NNSE}: We reiterate that finding latent factors that are interpretable is not just a luxury but a bare necessity in our setting as we need to understand what kind of latent semantic factors play a role in maximizing response rate. For this, we compared the latent factors induced by LDA and NNSE, examples of which are in \reftbl{tab:lda-vs-nnse}. In this table, four randomly selected latent factors induced each by LDA and NNSE are shown. Also, for each latent factor, top two most active questions in that dimension are shown. For easy reference, the main theme of each question is  manually marked in bold. From this table, we observe that NNSE is able to produce much more interpretable latent semantic factors compared LDA. Such lack of interpretability in LDA topics was also observed in another prior work  \cite{althoff2014ask}. Given the interpretability advantage with NNSE, we use the latent factors induced by this method in subsequent analysis.



Having successfully induced interpretable semantic factors using NNSE which have good number of questions attached to them, we analyzed the dimensions of questions with extremely high and extremely low reply rates. Please note that such latent factors are induced separately for each domain. Experimental results comparing NNSE latent factors in three domains, overall response rate in the domain, response rate over questions in the factor, and examples of top questions in each such factor are shown in \reftbl{tab:sem_factors}. Based on this table, we list below a few trends. We point out that this analysis and trend recognition would have been impossible without the ability to automatically induce interpretable semantic factors.

\subsubsection{Actors}
We found that adulation techniques worked well in eliciting a response for actors: 15.88\% response rate in Actor latent factor 524 in \reftbl{tab:sem_factors} compared to domain response rate of 5.19\%. Based on the top questions in this factor, we can easily identify that this is a fan-related factor. Authors seem to reply more if the inquirer describes himself as a huge fan or if he expresses some liking for their movies and role. We also learnt that actors weren't very comfortable when it came to questions diving into their non-camera life (Actor factor 880). Also many actors were evasive when asked about their favorite actors, movies, meals etc (Author factor 852).

\subsubsection{Politicians}
We observe that Politicians were prompt in clarifying all fund related issues pertaining to their campaigns (Politician factor 927 in \reftbl{tab:sem_factors}). Whereas not many politicians seemed to be happy in taking questions on wage rise and the job situations in the country (Politician factor 304).

\subsubsection{Author}
We observe that many users inquired authors about how they can pursue a career in writing, even more asked about writing advices. We found that such questions were generously replied: 36.53\% response rate in factor 742 of the Author domain, compared to domain response rate of 17.62\%. Also, authors answered a lot of questions that questioned about their ideas, thoughts and preferences (Author factor 136). However, they were a little less responsive when asked about inspiration (factor 4) or favorites (factor 118). This might be attributed to the fact that questions of these types are extremely frequently posed to authors, and due to the redundancy, they may answer only a few of them (please note that the response rate in these factors are not 0).

\subsection{Summary of Results}
From \refsec{sec:predict_response}, we observe that all our designed factors in conjunction beat the Random and the Bag-of-words baseline for all the domains. We also use far less features compared to the thousands of features in BoW (Unigram). This clearly demonstrates that we have arrived at a good mix of concise factors that are helpful in understanding response rate.

From \refsec{sec:sem_factors}, we see that our technique was able to capture some hard to find semantic factors that resulted in high reply rates. This also allowed us to identify factors in questions that are scantily replied.

\section{Conclusion}
\label{sec:conclusion}

Question-Answering forms an integral part of our everyday communication. While some questions elicit a lot of responses, many others go unanswered. In this paper, we present a large-scale empirical analysis to identify factors underlying response-eliciting questions. To the best of our knowledge, this is the first such analysis of its kind. In particular, we focus on the \misrfull (\misr{}) online setting where there are multiple users  asking questions to a single responder, and where
the responder has a choice to not answer any particular question. We used a novel dataset from the website Reddit.com, and considered several factors underlying questions, viz.,  orthographic, temporal, syntactic, and semantic. For semantic features, we used a sparse non-negative matrix factorization technique to automatically identify \emph{interpretable} latent factors. Because of this automated analysis, we are able to observe a few interesting and non-trivial trends. 
For instance we observed that all the advice related questions were generously entertained by Authors, as long as they carried some context about their writing pursuits. Similarly Actors were keen on making people aware about the behind-the-scene events, whenever asked. These trends are hard to capture otherwise, as designing a system to detect such particular cases requires training over large annotated corpus.  

As part of future work, we hope to explore other factorization techniques, e.g., hierarchical latent factors, for even more effective and interpretable latent factors. Additionally, we hope to use the insights gained in this study to explore how an existing question may be rewritten to elicit response from voluntary responders. We hope to make all the datasets and code publicly available upon publication of the paper.

\section{Acknowledgements}

This research is supported in part by gifts from Google Research and Accenture Technology Labs.



%
\bibliographystyle{IEEEtran}
\bibliography{asking_good_questions}  
\end{document}